\documentclass{article}

\usepackage[pdftex]{graphicx}
\usepackage{booktabs} 
\usepackage{bm}
\usepackage{amsmath}
\usepackage{amssymb}
\usepackage{multirow}
\usepackage{threeparttable}
\usepackage{verbatim}

\usepackage{epstopdf}
\usepackage{tikz-cd}

\usepackage{times}
\newtheorem{definition}{Definition}

\newcommand{\vc}[1]{\mathbf{#1}}

\usepackage[caption=false]{subfig}
\usepackage{caption}
\usepackage{algpseudocode}

\newcommand{\TheName}{\texttt{\bf RIFLE}}
\newcommand{\name}{\TheName}
\newcommand{\namea}{\TheName-A}
\newcommand{\nameb}{\TheName-B}

\usepackage{hyperref}

\usepackage[accepted]{icml2020}

\icmltitlerunning{RIFLE: Backpropagation in Depth for Deep Transfer Learning through Re-Initializing the Fully-connected LayEr}

\begin{document}

\twocolumn[
\icmltitle{\TheName: Backpropagation in Depth for Deep Transfer Learning through \underline{R}e-\underline{I}nitializing the \underline{F}ully-connected \underline{L}ay\underline{E}r   }



\icmlsetsymbol{equal}{*}

\begin{icmlauthorlist}
\icmlauthor{Xingjian Li}{equal,baidu,um}
\icmlauthor{Haoyi Xiong}{equal,baidu}
\icmlauthor{Haozhe An}{baidu}
\icmlauthor{Chengzhong Xu}{um,umlab}
\icmlauthor{Dejing Dou}{baidu}
\end{icmlauthorlist}

\icmlaffiliation{baidu}{Big Data Lab, Baidu Research, Beijing, China}
\icmlaffiliation{um}{Faculty of Science and Technology, University of Macau, Macau SAR, China}
\icmlaffiliation{umlab}{State Key Lab of IOTSC, Department of Computer Science, University of Macau, Macau SAR, China}

\icmlcorrespondingauthor{Xingjian Li, Haoyi Xiong}{\{lixingjian,xionghaoyi\}@baidu.com}


\icmlkeywords{Transfer Learning, Deep Learning, Machine Learning, ICML}

\vskip 0.3in
]



\printAffiliationsAndNotice{\icmlEqualContribution} 
\begin{abstract}
   Fine-tuning the deep convolution neural network (CNN) using a pre--trained model helps transfer knowledge learned from larger datasets to the target task. While the accuracy could be largely improved even when the training dataset is small, the transfer learning outcome is usually constrained by the pre-trained model with close CNN weights~\cite{liu2019towards}, as the backpropagation here brings smaller updates to deeper CNN layers. In this work, we propose \TheName-- a simple yet effective strategy that deepens backpropagation in transfer learning settings, through periodically \underline{R}e-\underline{I}nitializing the \underline{F}ully-connected \underline{L}ay\underline{E}r with random scratch during the fine-tuning procedure. \TheName\ brings meaningful updates to the weights of deep CNN layers and improves low-level feature learning, while the effects of randomization can be easily converged throughout the overall learning procedure. The experiments show that the use of \TheName\ significantly improves deep transfer learning accuracy on a wide range of datasets, outperforming known tricks for the similar purpose, such as Dropout, DropConnect, Stochastic Depth, Disturb Label and Cyclic Learning Rate, under the same settings with 0.5\%--2\% higher testing accuracy. Empirical cases and ablation studies further indicate \TheName~ brings meaningful updates to deep CNN layers with accuracy improved.
   \end{abstract}

\section{Introduction}
Blessed by tons of labeled datasets that are open to the public, deep learning~\cite{lecun2015deep} has shown enormous success for image classification and object detection tasks in the past few years. In addition to training a deep convolution neural network (CNN) from scratch with empty or random initialization, a more effective way might be fine-tuning a deep CNN using the weights of a pre--trained model, e.g., ResNet-50 trained using ImageNet, as the starting point~\cite{kornblith2019better}, since the learning dynamics would converge to a local minima of loss that is close to its starting point~\cite{neyshabur2015path,soudry2018implicit,liu2019towards}. To the end, such deep transfer learning paradigms reuse convolutional filters of well-trained models~\cite{yosinski2014transferable,long2015learning}, and are capable of extracting useful features from the target dataset for better classification after a short fine-tuning procedure. 

While above fine-tuning practice greatly enhances the accuracy of image classification even when the training dataset is small~\cite{bengio2012deep,kornblith2019better}, the room for further enhancement still exists. Due to the use of well-trained filters, the fine-tuning procedure frequently converges very quickly, while most of CNN weights have not been updated sufficiently. Although the fully-connected (FC) layer would be fine tuned with superior performance to classify the training datasets~\cite{zhang2017defense}, the features learned for classifications might still be largely based on the source datasets of transfer learning rather than the target one~\cite{yosinski2014transferable,long2015learning}. From the error backpropagation's perspectives~\cite{rumelhart1988learning,lecun1988theoretical}, the fast convergence of the FC layer leads to trivial weight updates to CNN~\cite{liu2019towards}, while the updates would become smaller and smaller with the depth of CNN layers~\cite{srivastava2015training}. 
In such way, compared to the CNN weights learned from the target training dataset directly, the fine-tuning procedure seldom changes or improves the deep CNN layers~\cite{liu2019towards}.
Our research hereby assumes that the in-depth backpropagation could further improve the performance of transfer learning based on pre--trained models.

To achieve the goal, tons of tricks, including Dropout~\cite{gal2016dropout}, Dropconnect~\cite{wan2013regularization}, Stochastic depth~\cite{huang2016deep}, Cyclic learning rate~\cite{smith2017cyclical}, Disturb label~\cite{xie2016disturblabel} and etc., have been proposed in supervised learning settings, especially to train very deep neural networks~\cite{srivastava2015training}. Our research, however, finds these efforts still might not work in-depth for transfer learning, as the FC layer converges so fast and overfits to the target training dataset with features extracted by the pre--trained CNN filers~\cite{li2019delta}.  We thus aim at proposing a novel backpropagation strategy that appropriately updates deep CNN layers, in transfer learning settings, against the fast convergence of FC. 

In this way, we propose a simple yet effective strategy, namely \TheName, which makes backpropagation in-depth for deep transfer learning through \emph{\underline{R}e-\underline{I}nitializing the \underline{F}ully-connected \underline{L}ay\underline{E}r}. 
More specifically, we equally divide the whole fine-tuning epochs into several periods, and re-initialize weights of the fully connected layer with random weights at the beginning of each period. Furthermore, cyclic learning rate in the same pace has been used to update the FC layer to ensure the final convergence. To verify the performance of \TheName, we carry out extensive experiments based on various popular image classification datasets with diverse types of visual objects. \TheName\ stably improve the fine-tuning procedures and outperforms the aforementioned strategies under the same settings.
In summary, our contributions are as follows:
\begin{itemize}
\item This work is the first attempt to improve deep transfer learning through pushing meaningful updates to the deep CNN layers, through the error backpropagation with the target training datasets. In a micro perspective, \TheName\ re-initializes the FC layer with random weights and enlarges the fitting error for deeper backpropagation. From a macro perspective, \TheName\ periodically incorporates non-trivial perturbation to the fine-tuning procedures and helps training procedure to escape the ``attraction'' of pre--trained CNN weights~\cite{zhou2019toward}. Furthermore, \TheName\ is designed to randomly re-initialized the FC layer only and might not hurt the convergence of overall CNN.

\item The existing regularizers for deep transfer learning, such as $L^2$ (weight decay)~\cite{bengio2012deep}, L$^2$-SP~\cite{li2018explicit}), and DELTA~\cite{li2019delta}, intend to improve the knowledge transfer through ensuring the consistency between the pre--trained model and the target one from weights and/or parameter perspectives. Our research, however, assumes such regularization reduces the effect of error backpropagation and might bring less meaningful modifications to the network, especially the deep CNN layers. On the other hand, when such regularization for knowledge transfer is disabled or weakened, the deep transfer learning algorithm might overfit to the training dataset with poor generalization performance. To balance the these two issues, \TheName\ is proposed. \TheName\ avoids overfitting through incorporating additional noise by random re-initialization while improving the knowledge transfer through backpropagation in-depth.


\item Through extensive experiments covering various transfer learning datasets for image classification, we demonstrate that, on top of two deep transfer learning paradigms (i.e., the vanilla fine-tuning and the one with L$^2$-SP~\cite{li2018explicit}), \TheName\ could achieve stable accuracy improvement with 0.5\%-2\% higher testing accuracy. It also outperforms existing strategies, including Dropout~\cite{gal2016dropout} (on both CNN layers and FC layers), Dropconnect~\cite{wan2013regularization}, Stochastic depth~\cite{huang2016deep}, Cyclic learning rate~\cite{smith2017cyclical} and Disturb label~\cite{xie2016disturblabel}, which have been proposed for similar purposes, under the same settings. We also conducted extensive empirical studies and an ablation study to show that (1) \TheName\ can make backpropagation in-depth and bring significant modifications to the weights of deeper layer; and (2) the weight modification brought by \TheName\ benefits to the generalization performance through enhancing the capacity of lower-level feature learning.
\end{itemize}
In addition to above contributions, to the best of our knowledge, \TheName\ is also yet the first algorithmic regularization~\cite{allen2019learning} designed for deep transfer learning purposes through weights randomization. \TheName\ can work together with an explicit deep transfer learning regularizer, such as~\cite{li2018explicit}, and makes a complementary contribution by further improving the generalization performance.

\section{Related Work}
In this section, we first introduce the related work of deep transfer learning with the most relevant work to our study.

\subsection{Deep Transfer Learning in General}
Transfer learning refers to a type of machine learning paradigms that aim at transferring knowledge obtained in the source task to a (maybe irrelevant) target task~\cite{pan2010survey,caruana1997multitask}, where the source and target tasks can share either the same or different label spaces. In our research, we primarily consider the inductive transfer learning with a different target label space for deep neural networks. As early as in 2014, authors in ~\cite{donahue2014decaf} reported their observation of significant performance improvement through directly reusing weights of the pre-trained source network to the target task, when training a large CNN with a tremendous number of filters and parameters. 

However, in the interim, while reusing all pre-trained weights, the target network might be overloaded by learning tons of inappropriate features (that cannot be used for classification in the target task), while the key features of the target task have been probably ignored. In this way, Yosinki \emph{et al.}~\cite{yosinski2014transferable} proposed to understand whether a feature can be transferred to the target network, through quantifying the ``transferability'' of features from each layer considering the performance gain. Furthermore, Huh \emph{et al.}~\cite{huh2016makes} made an empirical study on analyzing features that CNN learned from the ImageNet dataset to other computer vision tasks, so as to detail the factors affecting deep transfer learning accuracy.~\cite{ge2017cvpr} use a framework of multi-task learning by incorporating examples from the source domain which are similar to examples of the target task. In recent days, this line of research has been further developed with an increasing number of algorithms and tools that can improve the performance of deep transfer learning, including subset selection ~\cite{ge2017cvpr,cui2018large}, sparse transfer \cite{liu2017sparse}, filter distribution constraining \cite{aygun2017exploiting}, and parameter transfer ~\cite{zhang2018parameter}. 

\subsection{Regularization for Deep Transfer Learning}
Here, we review the knowledge transfer techniques that reuse pre-trained weights through the regularization. The square Euclidean distance between the weights of source and target networks is frequently used as the regularizer for deep transfer learning~\cite{li2018explicit}.  Specifically, ~\cite{li2018explicit} studied how to accelerate deep transfer learning while preventing fine-tuning from over-fitting, using a simple $L^2$-norm regularization on top of the ``Starting Point as a Reference'' optimization. Such method, namely $L^2$-SP, can significantly outperform a wide range of deep transfer learning algorithms, such as the standard $L^2$-norm regularization. Yet another way to regularize the deep transfer learning is ``knowledge distillation''~\cite{hinton2015distilling,romero2014fitnets}. In terms of methodologies, the knowledge distillation was originally proposed to compress deep neural networks~ \cite{hinton2015distilling,romero2014fitnets} through teacher-student network training, where the teacher and student networks are usually based on the same task~\cite{hinton2015distilling}. In terms of inductive transfer learning, authors in~\cite{yim2017gift} were first to investigate the possibility of using the distance of intermediate results (e.g., feature maps generated by the same layers) of source and target networks as the regularization term. Further,~\cite{zagoruyko2016paying} proposed to use the regularization of the divergence between activation maps 
for ``attention transfer''. 

In addition to the above explicit regularization, the implicit regularization or algorithmic regularization has been frequently discussed in deep learning settings~\cite{soudry2018implicit}. Generating and controlling the noise in the stochastic learning process, such as Dropout, cyclic learning rate and so on~\cite{gal2016dropout,huang2016deep,smith2017cyclical}, has been considered as a way to improve deep learning under the over-parameterization through implicit/algorithmic regularization. However, the performance improvement caused by these implicit/algorithmic regularizers has not been discussed in the context of deep transfer learning.

\subsection{Connection to Our Work} 
Compared to above work and other transfer learning studies, our work aims at providing a \emph{in-depth backpropagation} strategy that improves the performance of {deep transfer learning}. The intuition of \TheName\ is to periodically re-initialize the FC layer of CNN with random weights while ensuring the final convergence of the overall fine-tuning procedure with cyclical learning rate~\cite{smith2017cyclical} applied on the FC layer. In our work, we demonstrated the capacity of \TheName\ working with two deep transfer learning paradigms --- i.e., vanilla fine-tuning and the one with $L^2$-SP~\cite{li2018explicit} regularization, using a wide range of transfer learning tasks. The performance boosts with \TheName\ in all cases of experiments suggests that \TheName\ improves deep transfer learning with higher accuracy.

In terms of methodologies, the most relevant studies to our work are~\cite{hinton2012improving,wan2013regularization,huang2016deep,smith2017cyclical}, where all these methods intend to deepen the backpropagation through incorporating perturbation in the supervised learning procedure. More specific, Dropout~\cite{hinton2012improving} randomly omits parts of neurons of input or hidden layers during learning to regularize the backpropagation. Furthermore, DropConnect~\cite{wan2013regularization} generalizes the idea of Dropout by randomly dropping some connections between neurons. Stochastic depth~\cite{huang2016deep} incorporates layer-wise random shortcuts to train short networks with sufficient backpropagation while inferring over the complete one. Disturb label~\cite{xie2016disturblabel} randomly perturbs labels of the training batch in certain proportions and involves randomness in the loss layer. Furthermore, the use of cyclical learning rates~\cite{smith2017cyclical} incorporates randomness over the training procedure to escape shallow and sharp local minima while achieving decent convergence. Our methodology follows this line of research, where periodical random re-initialization of the FC layer makes backpropagation in depth while ensuring convergence of the overall fine-tuning procedure.

\section{Deep Transfer Learning with \TheName}
In this section, we first introduced the common algorithms used for deep transfer learning, then present the design of \TheName\ based on randomized regularization.

\subsection{Deep Transfer Learning with Regularization} 
The general deep transfer learning problem based on the pre--trained model is usually formulated as follow.

\begin{definition}[Deep Transfer Learning]
First of all, let's denote the training dataset for the desired task as $\mathbf{D}=\{(\vc{x}_1,y_1),(\vc{x}_2,y_2),(\vc{x}_3,y_3)\dots,(\vc{x}_n,y_n)\}$, where totally $n$ tuples are offered and each tuple $(\vc{x}_i,y_i)$ refers to the input image and its label in the dataset. 
We then denote $\mathbf{\omega}\in\ \mathbb{R}^{d}$ be the $d$-dimensional parameter vector containing all $d$ parameters of the target model. Further, given a pre-trained network with parameter $\mathbf{\omega_s}$ based on an extremely large dataset as the source, one can estimate the parameter of target network through the transfer learning paradigms. 
The optimization object with based deep transfer learning is to obtain the minimizer of $\mathcal{L}(\mathbf{\omega})$
\begin{equation} \label{eq:rtl}
\underset{w}{\mathrm{min}}\ \mathcal{L}(\omega)=\left\{ \frac{1}{n}\sum_{i=1}^n L(z( \vc{x}_{i}, \mathbf{\omega}), y_{i}) + \lambda\cdot\Omega(\mathbf{\omega},\mathbf{\omega}_s)\right\} \qquad
\end{equation}
where (i) the first term $\sum_{i=1}^n L(z( \vc{x}_{i}, \mathbf{\omega}), y_{i})$ refers to the empirical loss of data fitting while (ii) the second term $\Omega(\mathbf{\omega},\mathbf{\omega_s})$ characterizes the differences between the parameters of target and source network. The tuning parameter $\lambda >0\ $ balances the trade-off between the empirical loss and the regularization term.
\end{definition}
As was mentioned, two common deep transfer learning algorithms studied in this paper are vanilla fine-tuning~\cite{kornblith2019better} and $L^2$-SP~\cite{li2018explicit}. Specifically, these two algorithms can be implemented with the general based deep transfer learning with different regularizers

\subsubsection{Regularization for Fine-tuning}
The vanilla fine-tuning procedure incorporates a simple $L^2$-norm regularization (weight decay) of the weights to ensure the sparsity of weights, such that 
    \begin{equation}
        \Omega(\mathbf{\omega},\mathbf{\omega}_s) = \|\mathbf{\omega}\|_2^2.\label{eq:l2}
    \end{equation}
Note that such $L^2$ regularization is indeed independent with $\mathbf{\omega}_s$. Fine-tuning only adopts the weights of the pre-trained model $\mathbf{\omega}_s$ as the starting point of optimization, so as to transfer knowledge learned from source datasets.

\subsubsection{Regularization of L2-SP}
In terms of the regularizer, this algorithm\cite{li2018explicit} uses the squared-Euclidean distance between the target weights (i.e., optimization objective $\mathbf{\omega}$) and the pre-trained weights $\mathbf{\omega_s}$ of source network (listed in Eq~\ref{eq:l2sp}) to constrain the learning procedure.
    \begin{equation}
        \Omega(\mathbf{\omega},\mathbf{\omega}_s) = \|\mathbf{\omega} - \mathbf{\omega}_{s}\|_2^2\label{eq:l2sp}
    \end{equation}
In terms of optimization procedure, $L^2$-SP makes the learning procedure start from the pre-trained weights (i.e., using $\mathbf{\omega_s}$ to initialize the learning procedure).

In the rest of this work, we presented a strategy \TheName\ to improve the general form of deep transfer learning shown in Eq.~\ref{eq:rtl}, then evaluated and compared \TheName\ using the above two regularizers with common deep transfer learning benchmarks.

\begin{algorithm}[H]
    \caption{\TheName\ with SGD Optimizer}
    \begin{algorithmic}[1]
    \Procedure{\TheName}{$\mathcal{L},\mathbf{D},\mathbf{\omega_s}, P, T, \eta_\mathrm{max}$}
     \For{$t = 1, 2, 3..., T$}
     \State $\tau\gets t ~\mathbf{mod}~ P$
     \If{$\tau = 0$}
     \State \textcolor{blue}{/*Randomized Re-Initialization of FC*/}
     \State $\Omega_\mathrm{FC}\sim \mathcal{N}(0,\delta^2I)$
     \State $\Omega'_t\gets \mathrm{resetFC}(\Omega_t,\Omega_\mathrm{FC})$
     \Else
     \State $\Omega'_t\gets \Omega_t$
     \EndIf
     \State \textcolor{blue}{/*Cyclical Learning Rates*/}
     \State $\eta_t\gets\frac{1}{2}\cdot\eta_\mathrm{max}\cdot\mathrm{cosine}(2\pi\cdot\tau/{P})+\frac{1}{2}\cdot\eta_\mathrm{max}$
     \State $\mathbf{B}_t\gets\textbf{ mini-batch sampling from}~\mathbf{D}$
     \State \textcolor{blue}{/*Backpropagation with SGD Optimizer*/}
     \State $\Omega_{t+1}\gets\Omega_t-\eta_t\cdot \frac{1}{|\mathbf{B}_t|}\sum_{(x,y)\in\mathbf{B}_t}\nabla\mathcal{L}_{(x,y)}(\Omega'_t)$
     \EndFor
     \State \Return{$\Omega_T$}
    \EndProcedure
    \end{algorithmic}
    \label{alg:descending}
  \end{algorithm}

\subsection{\TheName\ Algorithm}
Given the regularized loss function $\mathcal{L}(\omega)$, weights of the pre-trained model $\Omega_s$, the target training dataset $\mathbf{D}=\{(\vc{x}_1,y_1),(\vc{x}_2,y_2),\dots,(\vc{x}_n,y_n)\}$, the number of iterations $P$ for the period of random re-initialization, the overall number of iterations for training $T$, and the maximal learning rate $\eta_\mathrm{max}$, we propose to use Algorithm~\ref{alg:descending} to train a deep neural network with \TheName\ with SGD Optimizer. Note that the optimizers (line 15) used for backpropagation are interchangeable with Adam, Momentum and etc., all based on the same settings of \TheName. The operator $\mathrm{resetFC}(\Omega_t,\Omega_{\mathrm{FC}})$ in line 7 refers to the operation that replaces the weights of the fully-connected layer in $\Omega_t$ with the random weights $\Omega_{\mathrm{FC}}$.

 Specifically, for each (e.g., the $t^{th}$) iteration of learning procedure, \TheName\ first checks whether a new period for random re-initialization of FC layer starts (i.e., whether $t~\mathbf{mod}~P = 0$). When a new period starts, \TheName\ draws a new random vector from the Gaussian distribution $\mathcal{N}(0,\delta^2I)$ as the weights of randomized FC layer $\Omega_\mathrm{FC}$, then reset the weights of the FC layer in the current model $\Omega_t$ with $\Omega_\mathrm{FC}$ to obtain $\Omega'_t$. Please refer to lines 4--8 for details. Later, \TheName\ adopts the simple cosine curve to adapt the cyclical learning rates~\cite{smith2017cyclical} then performs backpropagation based on $\Omega'_t$ and $\eta_t$ accordingly. Note that in Algorithm.~1, we present the design to incorporate the SGD optimizer for training. It would be appropriate to use other optimizers such Adam, Momentum and etc. that also incorporate $\Omega'_t$ and $\eta_t$ as inputs.

\subsubsection{Discussion} 
Note that \TheName\ strategy is derived from the stochastic gradient estimator used in stochastic gradient based learning algorithms, such as SGD, Momentum, conditioned SGD, Adam and so on. 
We consider \TheName\ as an alternative approach for descent direction estimation, where one can use a natural gradient-alike method to condition the descent direction or to adopt Momentum-alike acceleration methods on top of \TheName.
We are not intending to compare \TheName\ with any gradient-based learning algorithms, as the contributions are complementary. 
One can freely use \TheName\ to improve any gradient-based optimization algorithms (if applicable) with new descend directions.
\section{Experiment}
In this section, we present the experiments to evaluate \TheName\ with comparison to the existing methods.
\subsection{Datasets}
Eight popular transfer learning datasets are used to evaluate the effect of our algorithm.

\textbf{Stanford Dogs.} The Stanford Dogs dataset is used for the task of fine-grained image classification, containing images of 120 breeds of dogs. Each category contains 100 training examples. It provides bounding box annotations for the further purpose of vision specific algorithms. We use only classification labels during training. 

\textbf{Indoors 67.} Indoors 67 is a scene classification task containing 67 indoor scene categories, each of which consists of 80 images for training and 20 for testing. A major property unique to object recognition tasks is that, both spatial properties and object characters are expected to be extracted to obtain discriminative features.   

\textbf{Caltech-UCSD Birds-200-2011.} CUB-200-2011 contains 11,788 images of 200 bird species from around the world. Each species is associated with a Wikipedia article and is organized by scientific classification. This task is challenging because birds are small and they exhibit pose variations. Each image is annotated with bounding box, part location, and attribute labels though we do not use this information.

\textbf{Food-101.} Food-101 is a large scale data set consisting of 101 different kinds of foods. It contains more than 100k images. Each category contains 750 training examples in total. In this paper, we select two subsets for transfer learning evaluation, each containing 30 and 150 training examples per category, named Food-101-30 and Food-101-150 respectively. 

\textbf{Flower-102.} Flower-102 consists of 102 flower categories. 1020 images are used for training, 1020 for validation, and 6149 images for testing. Compared with other datasets, it is relative small since only 10 samples are provided for each class during training.

\textbf{Stanford Cars.} The Stanford Cars dataset contains 16,185 images of 196 classes of cars. The data is split into 8,144 training images and 8,041 testing images, where each class has been split roughly in a 50-50 split. Classes are typically at the level of Make, Model, Year, e.g. 2012 Tesla Model S or 2012 BMW M3 coupe.

\textbf{FGVC-Aircraft.} FGVC-Aircraft is composed of 102 different types of aircraft models, each of which contains 37 images for training and 37 for testing on average. It is a benchmark dataset for the fine grained visual categorization tasks. Bounding box annotations are not used for evaluation.

\textbf{Describable Textures Dataset.} Describable Textures Dataset (DTD) is a texture database, consisting of 5640 images. They are organized by a list of 47 categories, according to different perceptual properties of textures.

\setlength{\tabcolsep}{4pt}
\begin{table*}
\centering
\footnotesize
\caption{Comparison of regularization methods. ``Original'' refers to the performance of using the standard fine-tuning algorithm. ``DropConn.'' refers to DropConnection. ``CNN-DrO.'' refers to Dropout performed on CNN layers. ``StochasticD.'' refers to Stochastic Depth. ``CycleLR.'' refers to Cyclic Learning Rate. ``DisturbL.'' refers to Disturb Label. }  
\label{tab:mainres}
\begin{tabular}{|l|c|ccccccc|}\hline
\multicolumn{2}{|c|}{}&\multicolumn{7}{c|}{ Algorithm regularizers + $L^2$~\cite{bengio2012deep} }\\\hline
Datasets & Original & Dropout & {DropConn.} & {CNN{-}DrO.} & {StochasticD.} & CyclicLR. & DisturbL. & \TheName \\ \hline
CUB-200-2011 & 80.45$\pm$0.35  & 80.43$\pm$0.38  & 79.82$\pm$0.24  & 65.08$\pm$0.19 & 66.48$\pm$0.70 & 80.06$\pm$0.24 & 80.61$\pm$0.29 & \textbf{81.13$\pm$0.23} \\
FGVC-Aircraft & 76.58$\pm$0.36  & 76.52$\pm$0.58  & 75.76$\pm$0.71 & 74.83$\pm$0.37 & 60.94$\pm$0.74  & 75.63$\pm$0.43  & 76.65$\pm$0.33 & \textbf{77.82$\pm$0.41} \\
Flower-102 & 92.53$\pm$0.31  & 92.82$\pm$0.21  & 91.99$\pm$0.36  & 89.36$\pm$0.26  & 67.30$\pm$1.28 & 92.33$\pm$0.26  & 91.98$\pm$0.41 & \textbf{93.00$\pm$0.08} \\
DTD & 64.08$\pm$0.44 & 64.09$\pm$0.70 & 62.62$\pm$0.88  & 63.24$\pm$0.83 & 56.33$\pm$0.85  & 63.64$\pm$0.43 & \textbf{66.71$\pm$0.33} & 64.94$\pm$0.37 \\
Indoor-67 & 75.09$\pm$0.39  & 74.63$\pm$0.59  & 74.30$\pm$0.65  & 73.15$\pm$0.26  & 60.36$\pm$2.11  & 74.48$\pm$0.53  & 74.60$\pm$0.62 & \textbf{76.15$\pm$0.52} \\
Stanford Cars & 89.12$\pm$0.13  & 89.34$\pm$0.07  & 88.82$\pm$0.33  & 88.20$\pm$0.07  & 75.40$\pm$0.54  & 88.97$\pm$0.14  & 89.52$\pm$0.08 & \textbf{90.08$\pm$0.14} \\
Stanford Dogs & 79.21$\pm$0.18  & 79.24$\pm$0.15  & 78.92$\pm$0.32  & 78.07$\pm$0.26  & 67.97$\pm$1.04  & 78.74$\pm$0.15  & 79.76$\pm$0.10 & \textbf{80.10$\pm$0.23} \\
Food-101-30 & 60.38$\pm$0.15  & 60.09$\pm$0.23  & 58.85$\pm$0.28  & 57.41$\pm$0.32  & 42.40$\pm$0.79  & 60.57$\pm$0.21  & 58.02$\pm$0.15 & \textbf{62.31$\pm$0.27} \\
Food-101-150 & 75.81$\pm$0.14  & 75.71$\pm$0.19  & 75.45$\pm$0.15  & 74.53$\pm$0.29  & 64.11$\pm$0.33  & 75.31$\pm$0.05  & 75.35$\pm$0.08 & \textbf{76.36$\pm$0.16}\\\hline

\multicolumn{2}{|c|}{}&\multicolumn{7}{c}{ Algorithmic regularizers + $L^2$-$SP$~\cite{li2018explicit} }\\\hline
Datasets & Original & Dropout & {DropConn.} & {CNN{-}DrO.} & {StochasticD.} & CyclicLR. & DisturbL. & \TheName \\ \hline
CUB-200-2011 & 80.65$\pm$0.10  & 81.17$\pm$0.28  & 80.35$\pm$0.27  & 65.41$\pm$0.40  & 66.47$\pm$0.90  & 80.74$\pm$0.13 & 80.71$\pm$0.08	 & \textbf{81.73$\pm$0.20} \\
FGVC-Aircraft & 76.31$\pm$0.42  & 76.44$\pm$0.32  & 75.79$\pm$0.47  & 74.26$\pm$0.21 & 60.31$\pm$1.13  & 76.08$\pm$0.37 & 77.04$\pm$0.32 & \textbf{78.00$\pm$0.58} \\
Flower-102 & 91.98$\pm$0.34 & 92.10$\pm$0.25  & 91.86$\pm$0.32  & 89.09$\pm$0.43  & 68.95$\pm$0.93  & 91.91$\pm$0.16  & 91.79$\pm$0.48 & \textbf{92.46$\pm$0.18} \\
DTD & 69.41$\pm$0.47  & 69.21$\pm$0.25  & 68.92$\pm$0.45  & 63.40 $\pm$0.62 & 54.73$\pm$1.01  & 68.90$\pm$0.38  & 66.61$\pm$0.50 &\textbf{70.71$\pm$0.36} \\
Indoor-67 & 75.99$\pm$0.33  & 75.38$\pm$0.57  & 75.44$\pm$0.68  & 73.00$\pm$0.52  & 62.23$\pm$0.65  & 75.21$\pm$0.71  & 74.63$\pm$0.19 & \textbf{77.11$\pm$0.44} \\
Stanford Cars & 89.12$\pm$0.21  & 89.48$\pm$0.26  & 88.91$\pm$0.09  & 88.43$\pm$0.15  & 75.45$\pm$0.99  & 89.15$\pm$0.15  & 89.50$\pm$0.11 & \textbf{90.00$\pm$0.14} \\
Stanford Dogs & 88.32$\pm$0.14  & 88.73$\pm$0.14  & 87.93$\pm$0.14  & 78.47$\pm$0.16  & 67.94$\pm$0.65  & 88.07$\pm$0.13  & 79.96$\pm$0.21 & \textbf{88.96$\pm$0.06} \\
Food-101-30 & 61.25$\pm$0.23  & 61.01$\pm$0.24  & 59.66$\pm$0.54  & 57.73$\pm$0.27  & 42.10$\pm$1.15  & 61.06$\pm$0.15  & 57.98$\pm$0.32 &\textbf{62.84$\pm$0.25} \\
Food-101-150 & 77.02$\pm$0.15  & 77.04$\pm$0.16 & 76.62$\pm$0.18  & 73.98$\pm$0.15  & 65.28$\pm$0.61  & 76.52$\pm$0.11  & 75.35$\pm$0.17 & \textbf{77.36$\pm$0.04}\\\hline

\end{tabular}

\end{table*}

\subsection{Settings}
All experiments are implemented on the popular ResNet-50 architecture pre-trained with the ImageNet classification task. We use a batch size of 32. SGD with the momentum of 0.9 is used for optimizing all models. The initial learning rate is set to 0.01, using a cosine annealing learning rate policy. We train all models for 40 epochs. Normal data augmentation operations of random mirror and random crop are used for better performance in all experiments. Specifically, we first resize input images with the shorter edge being 256, in order to keep the original aspect ratio, following with normal data augmentation operations of random mirror and random crop to 224*224. Then inputs are normalized to zero mean for each channel. 

We compare several important regularization algorithms in transfer learning scenarios.  \emph{1)} Dropout and DropConnect on fully connected layers. For each method, we test the drop probability of 0.1, 0.2, 0.5 and report the best one. \emph{2)} Dropout on convolutional layers. We also test typical choices of 0.1, 0.2, 0.5 and report the best one. \emph{3)} Stochastic Depth. We use linear decay of the probability of skipping with stochastic depth for $p_0 = 1$ and $p_L = 0.5$,  following the best setting of \cite{huang2016deep}. Conceptually the first block which is the closest to the input is always active, and the last block which is the closest to the output is skipped with a probability of 0.5. \emph{4)} Cyclic learning rate. We divide the training procedure into $N$ cycles. We test choices of 2, 3, 4 and report the best one.  \emph{5)} Disturb label. We test the perturbation probability of 0.05, 0.1, 0.2 recommended by~\cite{xie2016disturblabel} and report the best result. We run each experiment five times and report the average top-1 accuracy. 

\subsection{Overall Comparisons}
We test our deep transfer learning algorithm with two common fine-tuning algorithms (regularization), which are standard fine-tuning ($L^2$) and using the starting point as reference ($L^2$-$SP$) in all cases. Through exhaustive experiments, we observe that:

\begin{itemize}
    \item \textbf{Comparison to the explicit regularizers for transfer learning.} \TheName~stably and significantly improves the performances of two common fine-tuning algorithms, including $L^2$ and $L^2$-$SP$, under all the cases. It demonstrates the feasibility of using algorithmic regularization \TheName\ to improve deep transfer learning on top of existing explicit regularizers.

    \item \textbf{Comparison to the algorithmic+explicit regularizers.}  \TheName~ outperforms the other state-of-the-art algorithms with similar purposes. These algorithms includes Dropout~\cite{gal2016dropout}, Drop Connect~\cite{wan2013regularization}, Stochastic Depth~\cite{huang2016deep}, Cyclic Learning Rate~\cite{smith2017cyclical} and Disturb Label~\cite{xie2016disturblabel}. For both two fine-tuning algorithms, \TheName~shows an improvement over all the existing methods on most datasets. We notice that CNN-Dropout and Stochastic Depth show poor performances on two fine-tuning algorithms. Among the compared algorithms, Dropout and Disturb label show relatively better performances, while performing less well than \TheName\ under most tasks. 
\end{itemize}
We can readily conclude that when \TheName+$L^2$ and \TheName+$L^2$-$SP$ significantly outperforms the vanilla explicit regularizers for deep transfer learning, while the simple combination of $L^2$ or $L^2$-$SP$ with other algorithmic regularizers, such as Dropout, Drop-connection and so on, could not perform as well as \TheName\ for deep transfer learning.  The comparison demonstrates the advancement achieved by \TheName.

\setlength{\tabcolsep}{6pt}
\begin{table}
\centering
\footnotesize
\caption{Comparison for Ablation Studies. \label{tab:ablres}}
\begin{tabular}{|l|ccc|}\hline
&\multicolumn{3}{c|}{$L^2$+\TheName\ and variants}\\\hline
Datasets & \name & \namea & \nameb \\ \hline
CUB-200-2011 & 81.13$\pm$0.23 & 81.07$\pm$0.29 & 79.84$\pm$0.28 \\
FGVC-Aircraft & 77.82$\pm$0.41 & 77.70$\pm$0.45 & 75.95$\pm$0.50 \\
Flower-102 & 93.00$\pm$0.08 & 92.17$\pm$0.19 & 92.44$\pm$0.16 \\
DTD & 64.94$\pm$0.37 & 65.12$\pm$0.53 & 64.11$\pm$0.84 \\
Indoor-67 & 76.15$\pm$0.52 & 75.94$\pm$0.11 & 74.02$\pm$0.38 \\
Stanford Cars & 90.08$\pm$0.14 & 89.88$\pm$0.20 & 88.65$\pm$0.19 \\
Stanford Dogs & 80.10$\pm$0.23 & 80.67$\pm$0.10 & 78.79$\pm$0.20 \\
Food-101-30 & 62.31$\pm$0.27 & 61.91$\pm$0.25 & 60.30$\pm$0.22 \\
Food-101-150 & 76.36$\pm$0.16 & 76.66$\pm$0.11 & 75.59$\pm$0.12 \\\hline
&\multicolumn{3}{c|}{$L^2$-$SP$+\TheName\ and variants}\\\hline
Datasets & \name & \namea & \nameb \\ \hline
CUB-200-2011 & 81.73$\pm$0.20 & 81.73$\pm$0.18 & 80.99$\pm$0.31 \\
FGVC-Aircraft & 78.00$\pm$0.58 & 77.91$\pm$0.38 & 75.89$\pm$0.49 \\
Flower-102 & 92.46$\pm$0.18 & 91.68$\pm$0.28 & 91.53$\pm$0.21 \\
DTD & 70.71$\pm$0.36 & 70.22$\pm$0.44 & 68.84$\pm$0.52 \\
Indoor-67 & 77.11$\pm$0.44 & 76.71$\pm$0.33 & 75.44$\pm$0.59 \\
Stanford Cars & 90.00$\pm$0.14 & 89.82$\pm$0.16 & 89.08$\pm$0.12 \\
Stanford Dogs & 88.96$\pm$0.06 & 88.90$\pm$0.11 & 88.00$\pm$0.07 \\
Food-101-30 & 62.84$\pm$0.25 & 62.48$\pm$0.09 & 61.19$\pm$0.24 \\
Food-101-150 & 77.36$\pm$0.04 & 77.96$\pm$0.08 & 76.44$\pm$0.11 \\\hline
\end{tabular}
\end{table}

\subsection{Ablation Study}
In order to investigate which is the most critical part in our algorithm, we do ablation study to show how each component affects the accuracy. First we define two modified versions of \TheName, named \namea\ and \nameb\ respectively. For \namea, we remove the cyclic learning rate component from our \TheName\ implementation and uses random re-initialization only. Therefore, the influence of different learning rates is cleared away in \namea. For \nameb, we remove the re-initialization component while using cyclic learning rates only. Thus, the only difference of \nameb\ compared with vanilla $L^2$ is the cyclic learning rate applied on the FC layer. We evaluate \namea\ and \nameb\ on all above tasks with both $L^2$ and $L^2$-$SP$ regularization. As observed in Table~\ref{tab:ablres}, \namea\ performs marginally worse than \name\ on most tasks, and is even superior to \name\ on particular datasets such as Food-101-150. \namea\ outperforms than \nameb and vanilla $L^2$ and $L^2$-$SP$. We can conclude that 
%
re-initializing the FC layer rather than cyclic learning rate is the founding brick of our algorithm.  

\subsection{Comparisons of Learning Curves} 
\begin{figure}[t]
\centering
\subfloat[$L^2$ (Train)]{\includegraphics[width=0.23\textwidth]{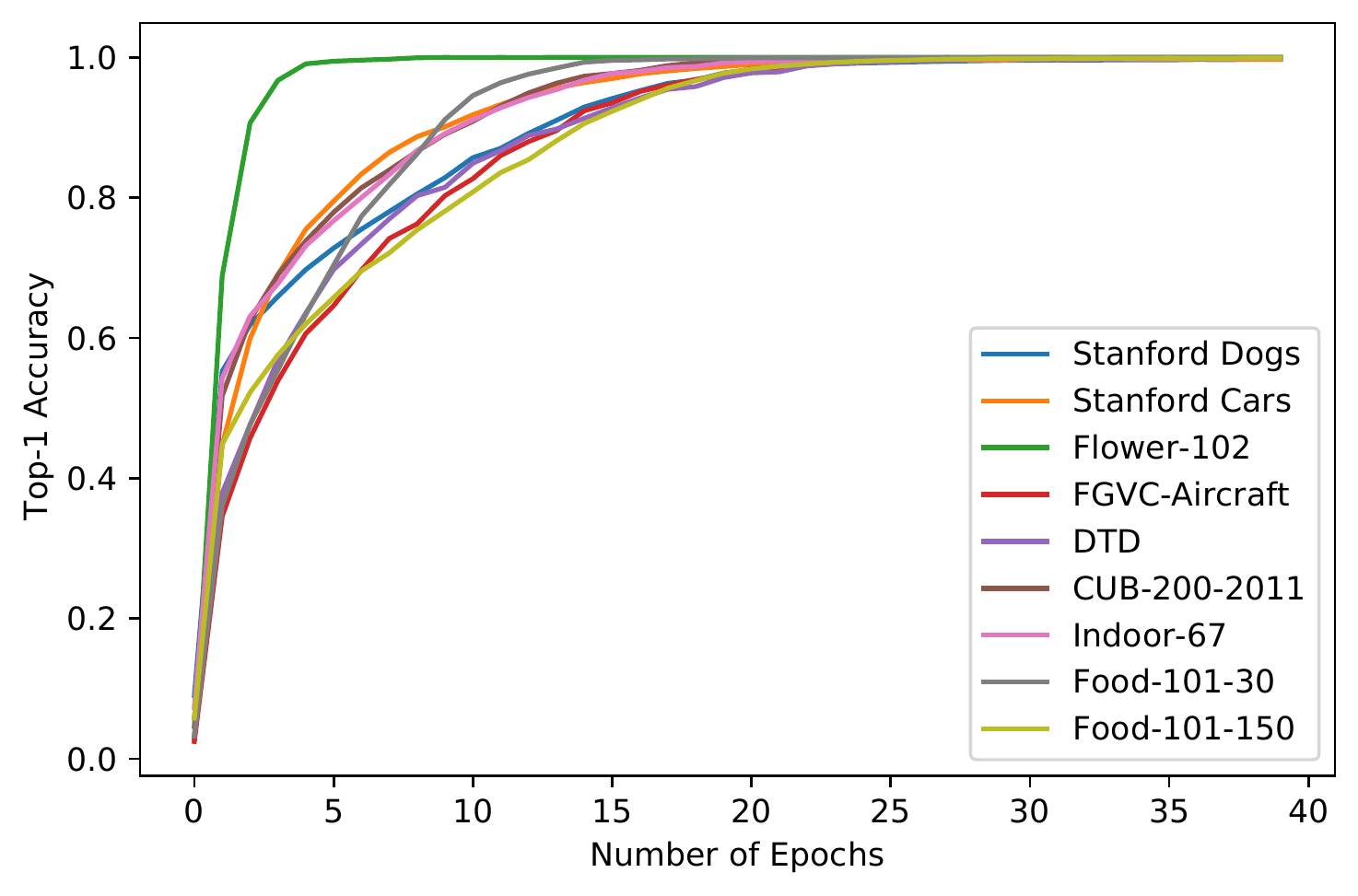} \label{sgd_train}}
\hfil
\subfloat[$L^2$ (Test)]{\includegraphics[width=0.23\textwidth]{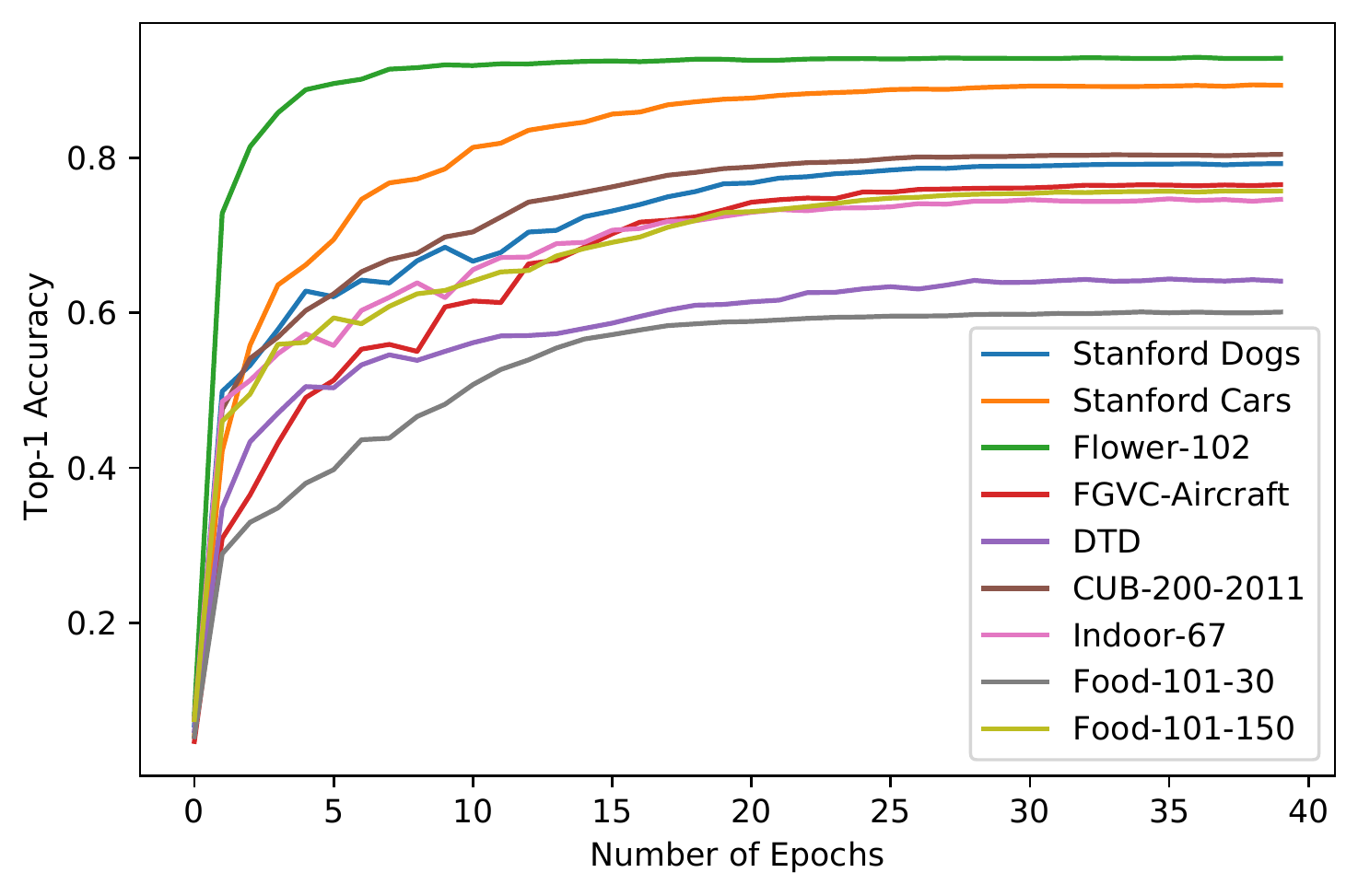} \label{sgd_test}}
\hfil
\subfloat[$L^2$+CyclicLR (Train)]{\includegraphics[width=0.23\textwidth]{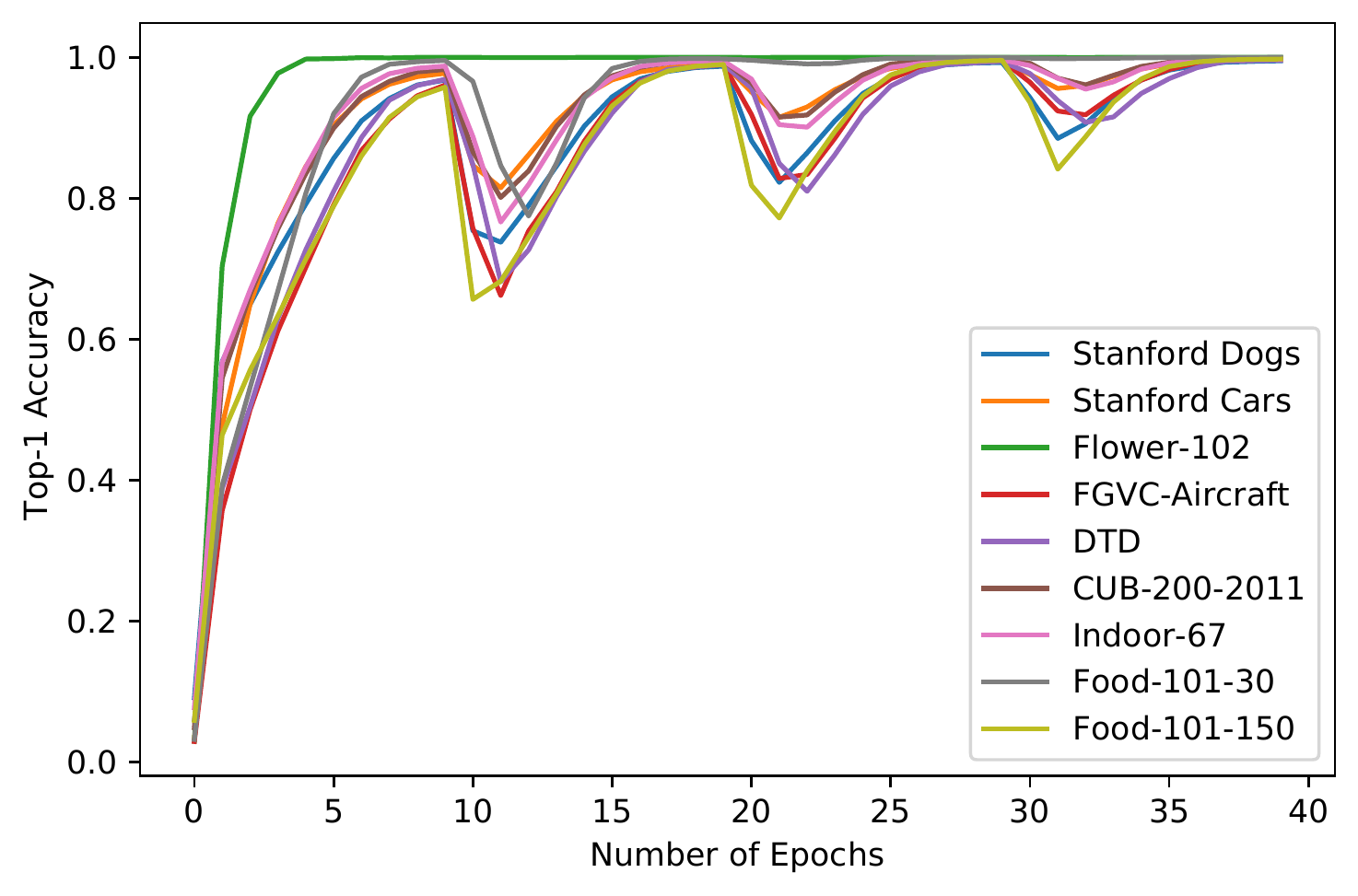} \label{cycliclr_train}}
\hfil
\subfloat[$L^2$+CyclicLR (Test)]{\includegraphics[width=0.23\textwidth]{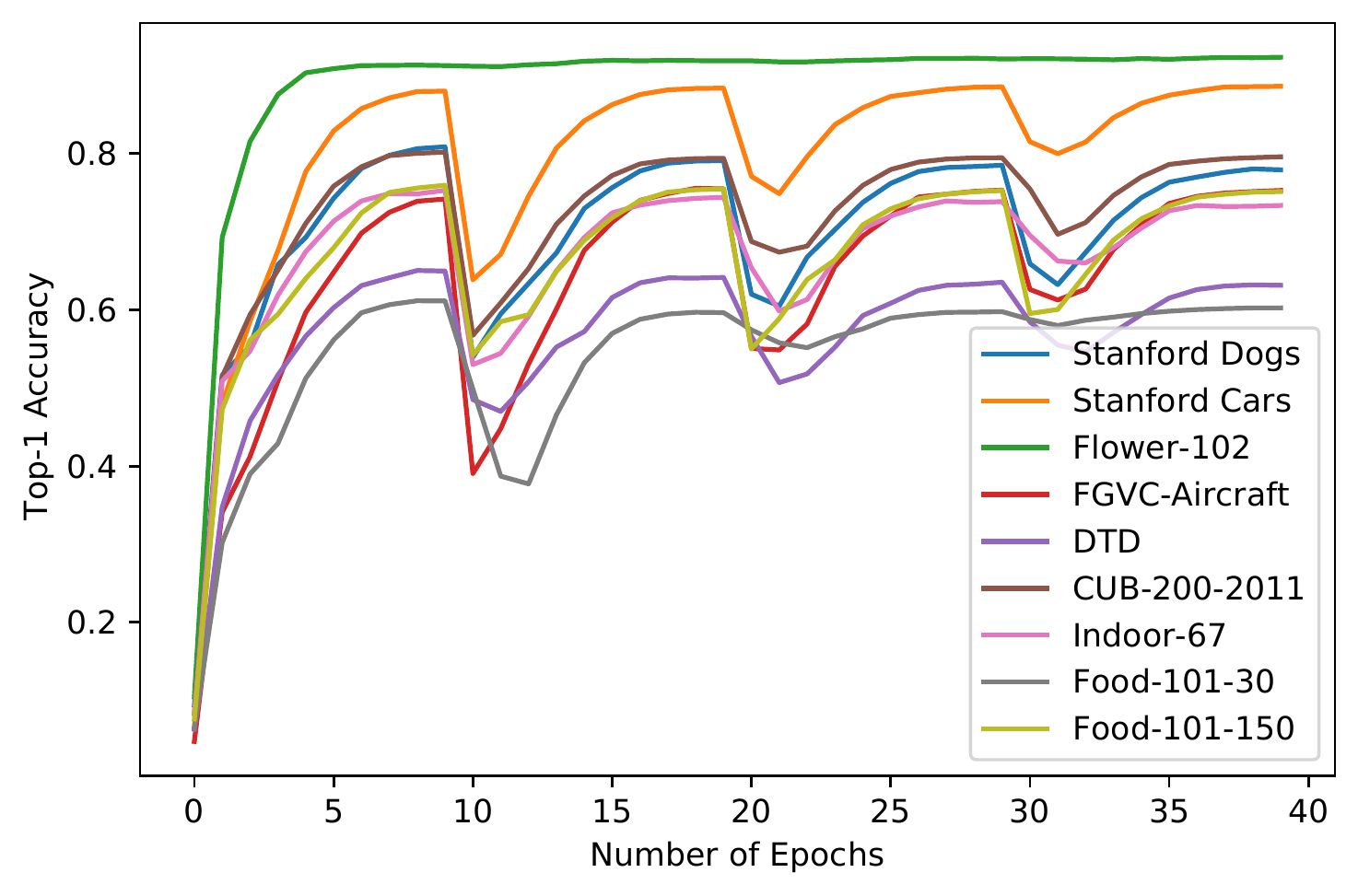} \label{cycliclr_test}}
\hfil
\subfloat[$L^2$+\namea\ (Train)]{\includegraphics[width=0.23\textwidth]{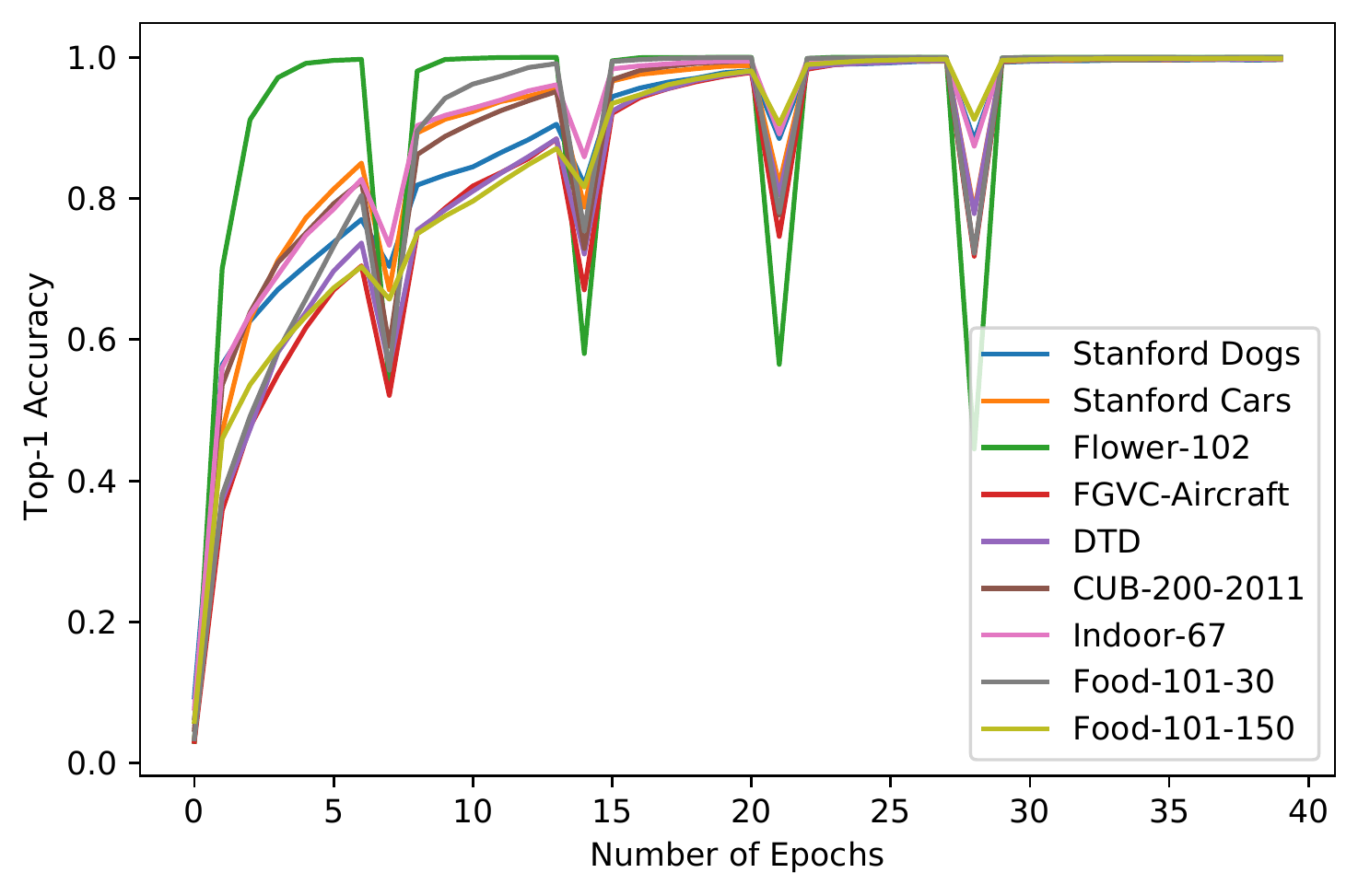} \label{rifle_train}}
\hfil
\subfloat[$L^2$+\namea\ (Test)]{\includegraphics[width=0.23\textwidth]{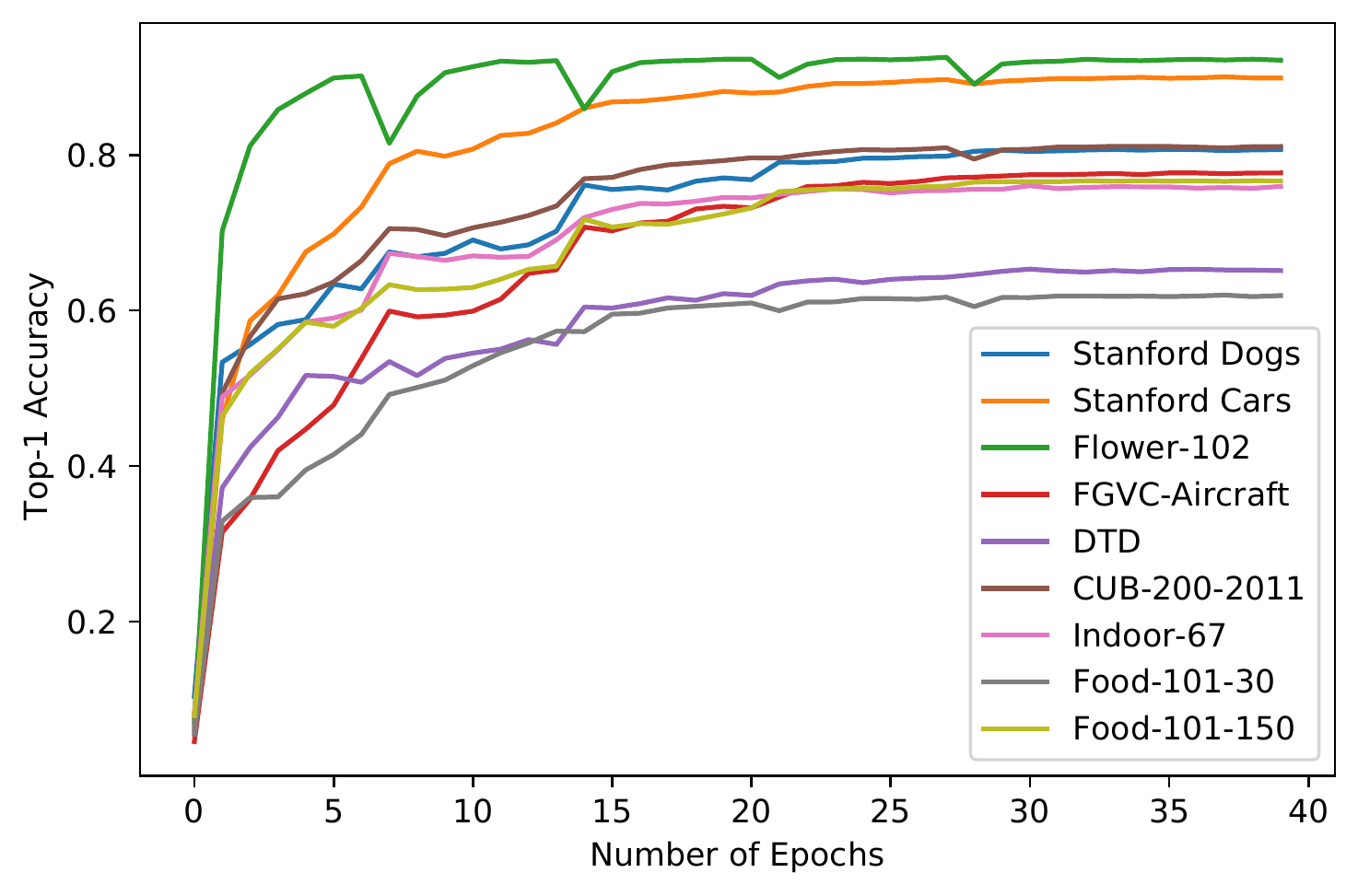} \label{rifle_test}}
\caption{Comparisons of Learning Curves.}
\label{fig:learning_curve}
\end{figure}

We further analyze the effect of \name\ by analysis of the learning curves, through the comparison with vanilla $L^2$ algorithm and the one based on $L^2$ with cyclic learning rate. As showed in Figure~\ref{sgd_train}, most tasks completely fit the training set after about half of the total number of epochs. For $L^2$ with cyclic learning rate, we find extremely sharp drops of the accuracy at the beginning of each period, due to the learning rate resetting. We notice in Figure~\ref{cycliclr_train} that, the magnitude of the decline diminishes as the number of training epochs increases. While for test set in Figure~\ref{cycliclr_test}, the accuracy always decreases severely at the first few epochs after learning rate resetting for most tasks. The training curve of \namea\ in Figure~\ref{rifle_train} is similar to the cyclic learning rate. However, compared to training curves in Figure~\ref{cycliclr_train}, the magnitude of the accuracy decline doesn't diminish in later training epochs for the one with cyclic learning rates. On the test set, \namea\ obtains much smoother accuracy curves compared to the one with the cyclic learning rate, which implies more stable generalization capacity. We infer that it is because the feature extractor part is perturbed by meaningful backpropagation brought by the re-initialized FC layer. 

We can conclude that, compared to cyclic learning rate or vanilla training procedure for deep transfer learning, \TheName\ brings more modifications to the weights during the learning procedure as the training curve is more unstable. Furthermore, such  modifications could improve the generalization performance of deep transfer learning, as the testing curve dominates in the comparison.


\subsection{Empirical Studies}

\begin{figure*}[t]
\centering
\subfloat[$L^2$ (CUB-200-2011)]{\includegraphics[width=0.24\textwidth]{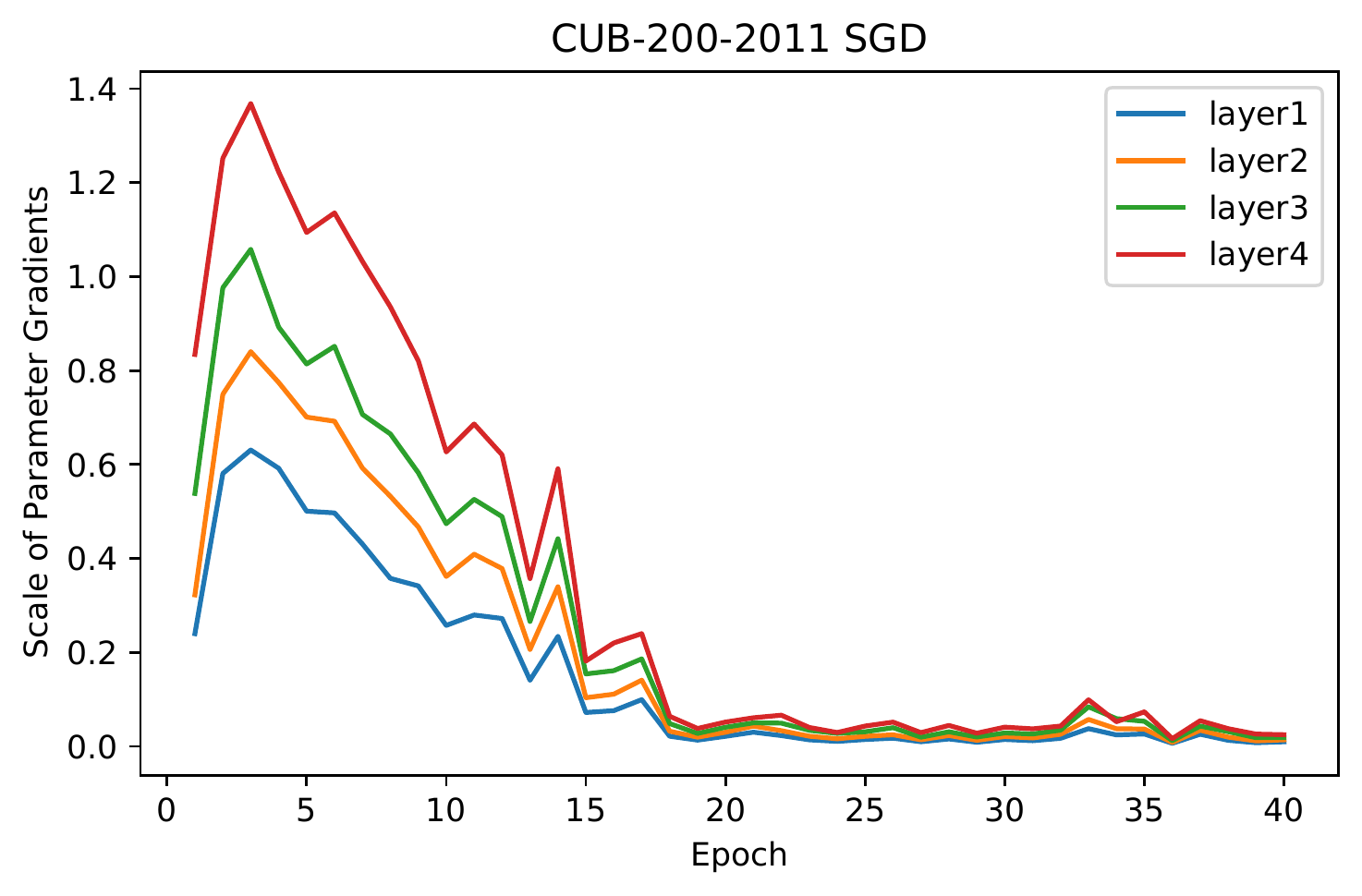} \label{cub_sgd}}
\hfil
\subfloat[$L^2$+\namea\ (CUB-200...)]{\includegraphics[width=0.24\textwidth]{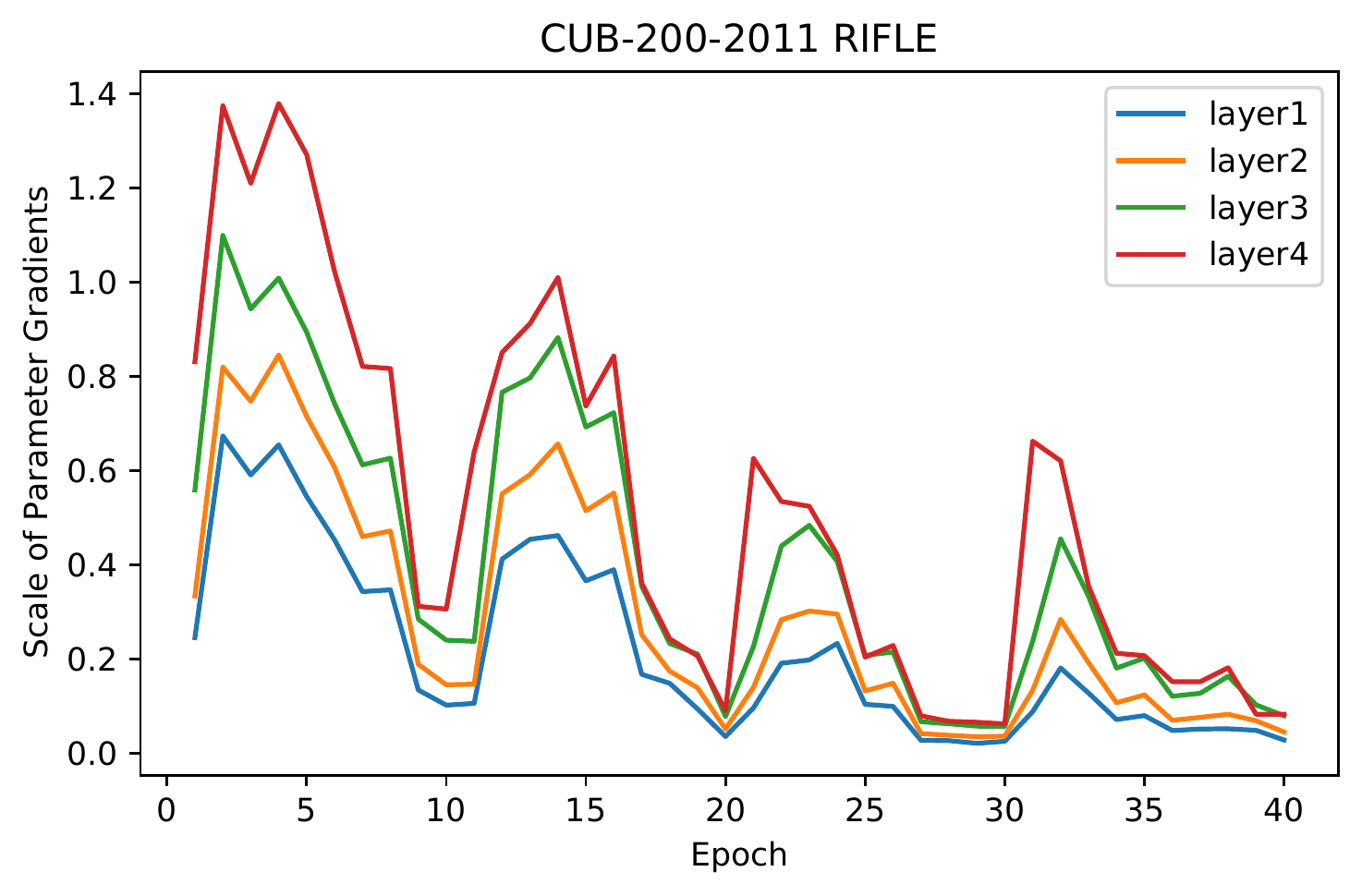} \label{cub_rifle}}
\hfil
\subfloat[$L^2$ (Flower-102)]{\includegraphics[width=0.24\textwidth]{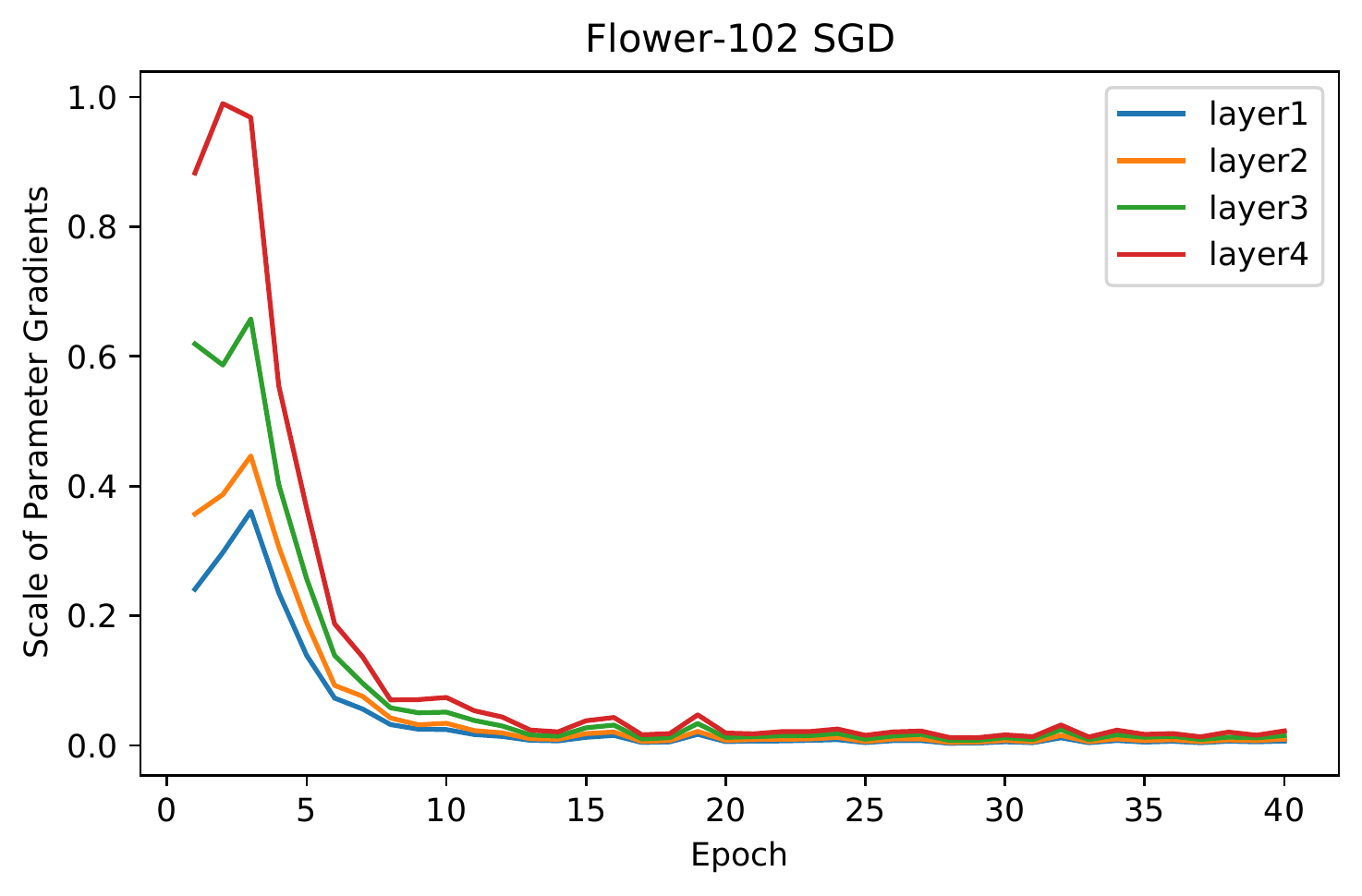} \label{flower_sgd}}
\hfil
\subfloat[$L^2$+\namea\ (Flower-102)]{\includegraphics[width=0.24\textwidth]{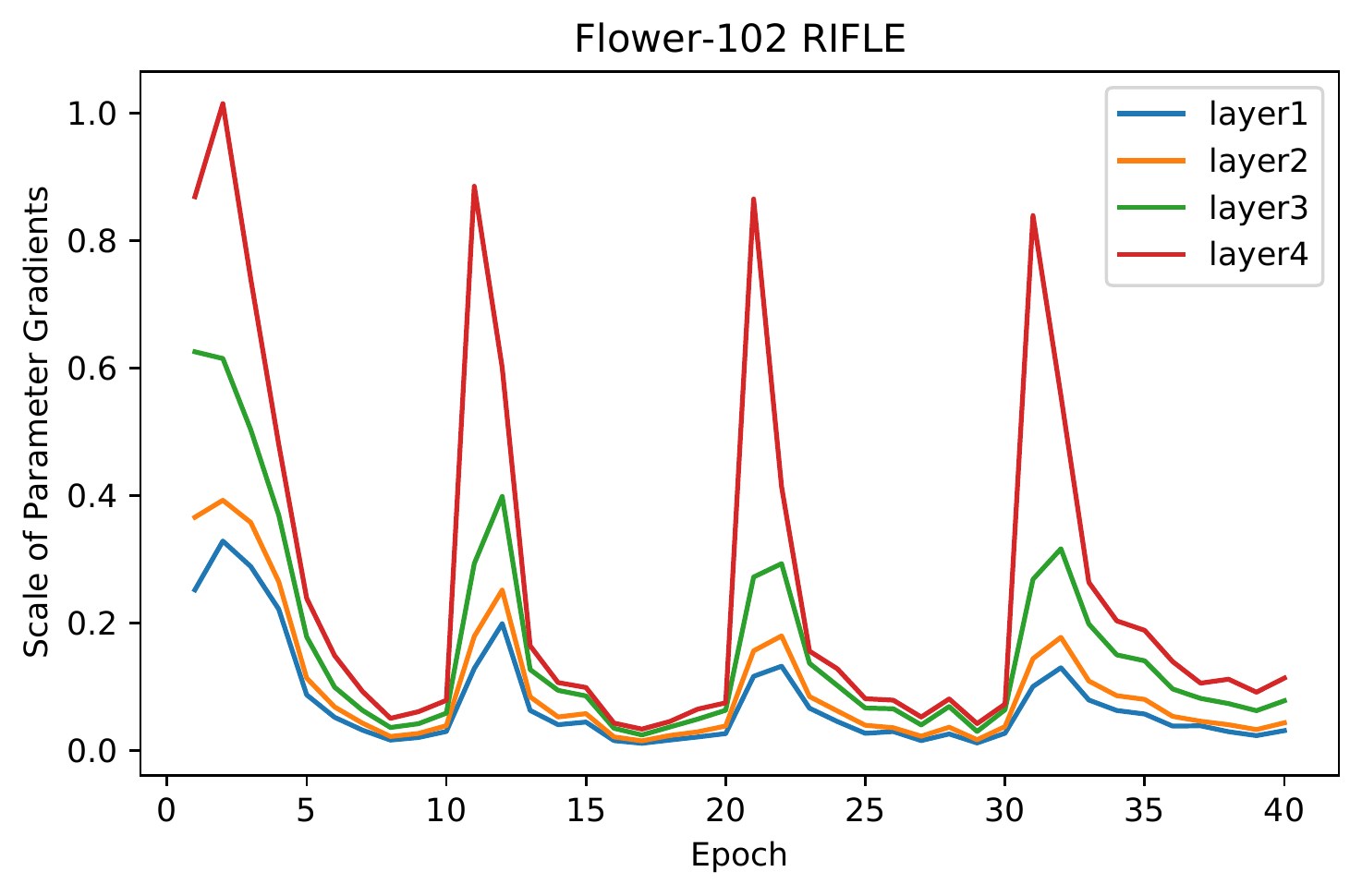} \label{flower_rifle}}
\caption{Scale of gradients corresponding to parameters of different layers.}
\label{fig:grad}
\vspace*{-2mm}
\end{figure*}

In this section, we intend to testify
    (1) Whether random re-initialization of the FC layer makes the backpropagation in-depth and updates the weights of deeper layers under transfer learning settings;
    (2) In which way, the random re-initialization of FC layer would improve the generalization performance of deep transfer learning.

\subsubsection{\TheName\ makes Backpropagation Deeper}

Our study finds \TheName\ makes backpropagation deeper and brings more modifications ton deeper layers.  In order to observe the phenomena, we measure the scale of gradients corresponding to different CNN layers during fine-tuning. To avoid the effects of cyclic learning rate, we perform the experiments using \namea\ with 4 times of re-initialization. We select the last $3\times3$ convolutional module in each of the four layers in ResNet-50. The Frobenius norm of the gradient of each parameter is calculated at the beginning of each training epoch. As illustrated in Figure~\ref{fig:grad}, we can observe that scales of gradients in vanilla $L^2$ decrease rapidly during training. For example, gradients of the CUB-200-2011 task almost vanish after 20 epochs when training with vanilla $L^2$ as showed in Figure~\ref{cub_sgd}. Flower-102 is even easier to convergence since gradients vanish after 10 epochs as showed in Figure~\ref{flower_sgd}. While for our proposed \namea, gradients are periodicity re-activated as demonstrated in Figure~\ref{cub_rifle} and~\ref{flower_rifle}, even though we use the exact same learning rate. We notice that parameters of layers close to the input (e.g. layer1) have smaller scales of gradients, which is consistent between vanilla $L^2$ and \namea.

\subsubsection{\TheName\ learns better low-level features}
Our study finds the low level features learned by \TheName\ are better than the vanilla deep transfer learning, through an empirical asymptotic analysis as follows.

\textbf{MLP Oracles.} We proposed to use two MLPs as the oracles to generate training datasets for source and target tasks respectively. The MLPs are both with two layers and with ReLU as activation. To make sure the knowledge transfer between two tasks, we make these two MLPs share the same first layer, such that $h_1(X)=W_2^\top\mathrm{ReLU}(W_1^\top X)$ and $h_2(X)=W_3^\top\mathrm{ReLU}(W_1^\top X)$. The input of networks $X$ is an 100-dimension vector (i.e., $X$ is a $100\times 1$ matrix), while the dimension of the hidden layer is 50 (i.e., $W_1$ is a $100\times 50$ random matrix and $W_2$ and $W_3$ are two $50\times 1$ random matrices). 

\textbf{Source/Target Tasks}. We collect the datasets for source and target tasks using $h_1(X)$ and $h_2(X)$ respectively, each of which is with 1000 samples drawn from standard Gaussian distributions. With $X_i$ randomly drawn from a 200-dimension Gaussian distribution as a training sample, we obtain $h_1(X_i)+\mathcal{N}(0,0.01)$ as the label of $x_i$ for the source task. With a similar approach, we also collect the datasets for the target task using $h_2(X)$.

\textbf{Transfer Learning}. We first use the generated source dataset to train an MLP with the same architecture from scratch through vanilla SGD via regression loss, and we obtain the teacher network model $h'_1(X)=W_2^{'\top}\mathrm{ReLU}(W_1^{'\top}X)$. Note that the weights $W'_2$ and $W'_1$ are so different from $W_2$ and $W_1$ as multiple solutions exist. With the weights $W'_2$ and $W'_1$ and the generated target dataset, we use $L^2$ and $L^2$+\TheName\ for transfer learning to obtain two networks $h'_2(X)=W_3^{'\top}\mathrm{ReLU}(W_1^{'\top} X)$ and $h^*_2(X)=W_3^{*\top}\mathrm{ReLU}(W_1^{*\top} X)$ respectively.

\textbf{Comparisons.} We find both $h'_2(X)$ and $h^*_2(X)$ achieves very good accuracy given the testing dataset generated by $h_2(X)$, while the testing loss of $h^*_2(X)$ (with MSE $3.98\times 10^{-3}$) is significantly lower than $h'_2(X)$ (with MSE $1.16\times 10^{-2}$). Furthermore, we also observe that $\mathrm{OT}(W^*_1,W_1)=0.1198\leq \mathrm{OT}(W'_1,W_1)=0.1397$, where $\mathrm{OT}(\cdot,\cdot)$ refers to the optimal transport distance~\cite{flamary2017pot} between the inputs. 
As the weights of deep layers are closer to the MLP oracles, we conclude that \TheName\ learns better low-level features.

\section{Conclusion}
 In this work, we propose \TheName-- a simple yet effective strategy that enables in-depth backpropagation in transfer learning settings, through periodically \underline{R}e-\underline{I}nitializing the \underline{F}ully-connected \underline{L}ay\underline{E}r with random scratch during the fine-tuning procedure.
 \TheName\ brings significant meaningful modifications to the deep layers of DNN and improves the low-level feature learning. The experiments show that the use of \TheName\ significantly improves the accuracy of deep transfer learning algorithms (based on explicit regularization) on a wide range of datasets. It outperforms known algorithmic regularization tricks such as dropout and so on, under the same settings. To the best of our knowledge, \TheName\ is yet the first algorithmic regularization method to improve the deep transfer learning algorithms based on explicit regularization such as~\cite{li2018explicit}. The contribution made in this work is complementary with the previous work.

\section{Acknowledgements}
This work is supported in part by National Key R\&D Program of China (No. 2019YFB2102100 to Chengzhong Xu, No. 2018YFB1402600 to Haoyi Xiong) and The Science and Technology Development Fund of Macau SAR (File no. 0015/2019/AKP to Chengzhong Xu). 


\bibliography{example_paper}
\bibliographystyle{icml2020}

\end{document}